\def\year{2015}

\documentclass[letterpaper]{article}
\usepackage{authblk}

\usepackage{aaai}
\usepackage{times}
\usepackage{helvet}
\usepackage{courier}

\usepackage{booktabs}
\usepackage{url}
\usepackage{latexsym}
\usepackage{amsmath}
\usepackage{amsfonts}
\usepackage{graphicx}
\usepackage{epstopdf}
\usepackage{multirow}
\usepackage{afterpage}

\usepackage{hyphenat}
\title{Building End-To-End Dialogue Systems \\ Using Generative Hierarchical Neural Network Models}

\author[*]{Iulian V. Serban}
\author[*]{Alessandro Sordoni}
\author[*]{Yoshua Bengio\thanks{Y.B. is a CIFAR senior Fellow}}
\author[*]{Aaron Courville}
\author[$\dag$]{Joelle Pineau}
\affil[*]{Department of Computer Science and Operations Research, Universit{\'e} de Montr{\'e}al, Montreal, Canada \protect\\ 
          \texttt{\{iulian.vlad.serban,alessandro.sordoni,yoshua.bengio,aaron.courville\} AT umontreal.ca}}
\affil[$\dag$]{School of Computer Science, McGill University, Montreal, Canada \hspace*{2mm} \texttt{jpineau AT cs.mcgill.ca}}

\date{}

\begin{document}

\makeatletter
\let\@fnsymbol\@arabic
\makeatother

\maketitle

\begin{abstract}
We investigate the task of building open domain, conversational dialogue systems based on large dialogue corpora using generative models. Generative models produce system responses that are autonomously generated word-by-word, opening up the possibility for realistic, flexible interactions.
In support of this goal, we extend the recently proposed hierarchical recurrent encoder-decoder neural network to the dialogue domain, and demonstrate that this model is competitive with state-of-the-art neural language models and backoff n-gram models. 
We investigate the limitations of this and similar approaches, and show how its performance can be improved by bootstrapping the learning from a larger question-answer pair corpus and from pretrained word embeddings.
\end{abstract}

\section{Introduction}
Dialogue systems, also known as interactive conversational agents, virtual agents and sometimes chatterbots, 
are used in a wide set of applications ranging from technical support services to language learning tools and entertainment \cite{young2013pomdp,shawar2007chatbots}. 
Dialogue systems can be divided into goal-driven systems, 
such as technical support services, and non-goal-driven systems, 
such as language learning tools or computer game characters. Our current work focuses on the second case, due to the availability of large corpora of this type, though the model may eventually prove useful for goal-driven systems also.

Perhaps the most successful approach to goal-driven systems has been to view the dialogue problem as a partially observable Markov decision process (POMDP) \cite{young2013pomdp}. 
Unfortunately, most deployed dialogue systems use hand-crafted features for the state and action space representations, and require either a large annotated task-specific corpus or a horde of human subjects willing to interact with the unfinished system. 
This not only makes it expensive and time-consuming to deploy a real dialogue system, but also limits its usage to a narrow domain.
Recent work has tried to push goal-driven systems towards learning with few examples using constraints on the POMDP \cite{gasic2013line} as well as learning the observed features themselves with neural network models \cite{henderson2014word}, yet such approaches still require either hand-crafted features or large corpora of annotated task-specific simulated conversations.


On the other end of the spectrum are the non-goal-driven systems \cite{ritter2011data,banchs2012iris,ameixa2014luke}. 
Most recently Sordoni et al.~\shortcite{sordoni2015aneural} and Shang et al.~\shortcite{shang2015neural} have drawn inspiration from the use of neural networks in natural language modeling and machine translation tasks~\cite{Cho-et-al-EMNLP2014-better}. 
There are several motivations for developing non-goal-driven systems. 
First, they may be deployed directly for tasks which do not naturally exhibit a directly measurable goal (e.g.\@ language learning) or simply for entertainment. Second, if they are trained on corpora related to the task of a goal-driven dialogue system (e.g.\@ corpora which cover conversations on similar topics) then these models can be used to train a user simulator, which can then train the POMDP models discussed earlier \cite{young2013pomdp,pietquin2013survey}. This would alleviate the expensive and time-consuming task of constructing a large-scale task-specific dialogue corpus. In addition to this, the features extracted from the non-goal-driven systems may be used to expand the state space representation of POMDP models \cite{singh2002optimizing}. This can help generalization to dialogues outside the annotated task-specific corpora.

Our contribution is in the direction of end-to-end trainable, non-goal-driven systems based on 
generative probabilistic models. 
We define the generative dialogue problem as modeling the utterances and interactive structure of
the dialogue. As such, we view our model as a cognitive system, which has to carry out natural language understanding,
reasoning, decision making and natural language generation in order to replicate or emulate the behavior of the agents
in the training corpus.
Our approach differs from previous work on learning dialogue systems through interaction with humans \cite{young2013pomdp,gasic2013line,cantrell2012tell,mohan2014learning}, because it learns off-line through examples of human-human dialogues and aims to emulate the dialogues in the training corpus instead of maximize a task-specific objective function. Contrary to explanation-based learning \cite{mohan2014learning} and rule-based inference systems \cite{langley2014abductive}, our model does not require a predefined state or action space representation. These representations are instead learned directly from the corpus examples together with inference mechanisms, which map dialogue utterances to dialogue states, and action generation mechanisms, which map dialogue states to dialogue acts and stochastically to response utterances.
We believe that training such a model end-to-end to minimize a single objective function, and with minimum reliance on hand-crafted features, will yield superior performance in the long run.
Furthermore, we focus on models which can be trained efficiently on large datasets 
and which are able to maintain state over long conversations.

We experiment with the well-established recurrent neural networks (RNN) and $n$-gram models. In particular, we adopt the hierarchical recurrent encoder-decoder (HRED)~\cite{sordoni2015ahier} and demonstrate that it is competitive with other models in the literature. We extend the model architecture to better suit the dialogue task. 
We show that performance can be substantially improved by bootstrapping from pretrained word embeddings and by pretraining the model on a larger question-answer pair (Q-A) corpus.
To carry out experiments, we introduce the \textit{MovieTriples} dialogue dataset based on movie scripts. 

\section{Related Work}
Modeling conversations on micro-blogging websites with generative probabilistic models was first proposed by Ritter et al.~\shortcite{ritter2011data}. They view the response generation problem as a translation problem, where a post needs to be translated into a response. Generating responses was found to be considerably more difficult than translating between languages, likely due to the wide range of plausible responses and lack of phrase alignment between the post and the response. 

Later, Shang et al.~\shortcite{shang2015neural} proposed to use the recurrent neural network framework for generating responses on micro-blogging websites. This was followed up by Sordoni et al.~\shortcite{sordoni2015aneural}, who extended the framework from status-reply pairs to triples of three consecutive utterances.

To the best of our knowledge, Banchs et al.~\shortcite{banchs2012iris} were the first to suggest using movie scripts to build dialogue systems. 
Conditioned on one or more utterances, their model searches a database of movie scripts and retrieves an appropriate response. 
This was later followed up by Ameixa et al.~\shortcite{ameixa2014luke}, who demonstrated that movie subtitles could be used to provide
responses to out-of-domain questions using an information retrieval system.


\section{Models}

We consider a dialogue as a sequence of $M$ utterances $D = \{ U_1, \ldots, U_M \}$ involving two interlocutors.
Each $U_m$ contains a sequence of $N_m$ tokens, i.e.  $U_m = \{w_{m, 1}, \ldots, w_{m, N_m}\}$, where $w_{m, n}$ is a random variable taking values in the vocabulary $V$ and representing the token at position $n$ in utterance $m$.
The tokens represent both words and \textit{speech acts}, e.g.\@ pause and end of turn tokens.
A generative model of dialogue parameterizes a probability distribution $P$ - governed by parameters $\theta$ - over the set of all possible dialogues of arbitrary lengths.
The probability of a dialogue $D$ can be decomposed:
\begin{equation}
{
\begin{split}
P_\theta(U_1, & \ldots, U_M) = \prod_{m = 1}^M P_\theta(U_m | U_{< m}), \\
& = \prod_{m = 1}^M \prod_{n = 1}^{N_m} P_\theta(w_{m, n} | w_{m, < n}, U_{< m}), \\
\end{split}}
\end{equation}
$U_{< m} = \{U_{1}, \ldots, U_{m-1}\}$, $w_{m, <n} = \{w_{m, 1}, \ldots, w_{m, n-1}\}$, i.e.~the tokens preceding $n$ in the utterance $U_m$. The task is analogous to language modeling, with the critical difference that \textit{speech acts} are included as separate tokens. Sampling from the model can be performed as in standard language modeling: sampling one word at a time from the conditional distribution $P_\theta(w_{m, n} | w_{m, < n}, U_{< m})$ conditioned on the previously sampled words. 

Using standard $n$-grams to compute joint probabilities over dialogues, e.g.\@ computing probability tables for each token given the $n$ preceding tokens, suffers from the curse of dimensionality and is intractable for any realistic vocabulary size.
To overcome this, Bengio et al.~\shortcite{Bengio-nnlm2003} proposed a distributed (dense) vector representation of words,  called \emph{word embeddings}, which parameterizes $P_\theta(w_{m, n} | w_{m, < n}, U_{< m})$ as a smooth function using a neural network.
By means of such distributed representations, the recurrent neural network (RNN) based language model~\cite{mikolov2010recurrent} has pushed state-of-the-art performance by learning long $n$-gram contexts while avoiding data sparsity issues.
Overall, RNNs have performed well on a variety of NLP tasks such as machine translation~\cite{Cho-et-al-EMNLP2014-better,sutskever2014sequence,bahdanau2014neural} and information retrieval~\cite{sordoni2015ahier}. 


\subsection{Recurrent Neural Network}

A recurrent neural network (RNN) models an input sequence of tokens $\{w_1, \dots, w_N\}$ using the recurrence:
\begin{align}
h_n &= f(h_{n-1}, w_n),
\vspace*{-1mm}
\end{align}
where $h_n \in \mathbb{R}^{d_h}$ is called a recurrent, or \emph{hidden}, state and acts as a vector representation of the tokens seen up to position $n$. 
In particular, the last state $h_N$ may be viewed as an order-sensitive compact summary of all the tokens. In language modeling tasks, the context information encoded in $h_n$ is used to predict the next token in the sentence:
\begin{equation*}
P_\theta(w_{n+1} = v | w_{\le n}) = \dfrac{\exp \left ( g(h_{n}, v) \right )}{\sum_{v'} \exp \left ( g(h_{n}, v') \right )}.
\end{equation*}
The functions $f$ and $g$ are typically defined as:
\begin{align}
f(h_{n-1}, w_n) & = \text{tanh} \left ( H h_{n-1} + I_{w_n} \right ), \\
g(h_n, v) & = O_{w_n}^{T} h_n,
\end{align}
The matrix $I \in \mathbb{R}^{d_h \times |V|}$ contains the input \emph{word embeddings}, i.e.~each column $I_j$ is a vector corresponding to token $j$ in the vocabulary $V$.
Due to the size of the model vocabulary $V$, it is common to approximate the $I$ matrix with a low-rank decomposition, i.e.~$I = X E$, where $X \in \mathbb{R}^{d_h \times d_e}$ and $E \in \mathbb{R}^{d_e \times |V|}$, and $d_e < d_h$. This approach has also the advantage that the embedding matrix $E$ may separately be bootstrapped (e.g. learned) from larger corpora. 
%
Analogously, the matrix $O \in \mathbb{R}^{d_h \times |V|}$ represents the output word embeddings, where each possible next token is projected into another dense vector and compared to the hidden state $h_n$. 
The probability of seeing token $v$ at position $n+1$ increases if its corresponding embedding vector $O_v$ is ``near'' the context vector $h_{n}$.
The parameter $H$ is called a \emph{recurrent} parameter, because it links $h_{n-1}$ to $h_{n}$.
%
All parameters are learned by maximizing the log-likelihood of the parameters on a training set using stochastic gradient descent.
\begin{figure*}
\centering
\includegraphics[natwidth=1186px,natheight=595px,scale=0.35]{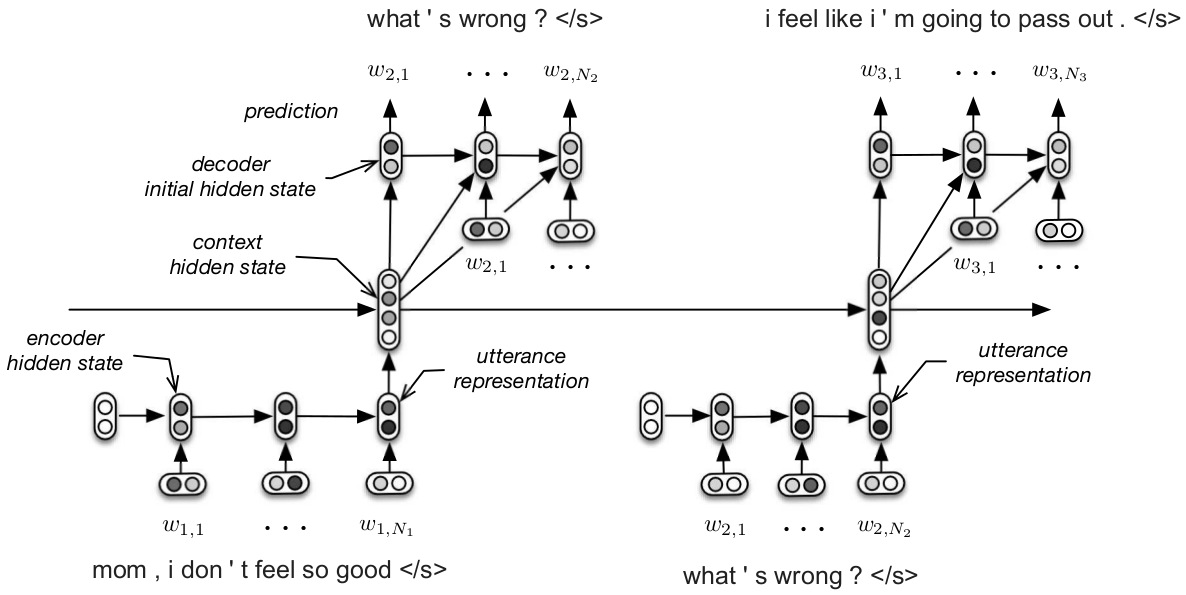}
\caption{\label{hrnn_model} The computational graph of the HRED architecture for a dialogue composed of three turns. Each utterance is encoded into a dense vector and then mapped into the dialogue context, which is used to decode (generate) the tokens in the next utterance. The encoder RNN encodes the tokens appearing within the utterance, and the context RNN encodes the temporal structure of the utterances appearing so far in the dialogue, allowing information and gradients to flow over longer time spans. The decoder predicts one token at a time using a RNN. Adapted from Sordoni et al.~\shortcite{sordoni2015ahier}.}
\end{figure*}


\subsection{Hierarchical Recurrent Encoder-Decoder}

Our work extends the hierarchical recurrent encoder-decoder
architecture (HRED) proposed by Sordoni et al.~\shortcite{sordoni2015ahier} for web query suggestion.
In the original framework, HRED predicts the next web query given the queries
already submitted by the user. The history of past submitted queries 
is considered as a sequence at two levels: a sequence of words for each web query and a sequence of queries.
HRED models this hierarchy of sequences with two RNNs: one at the word level and one at the query level. 
We make a similar assumption, namely, that a dialogue can be seen as a sequence of utterances which, in turn,
are sequences of tokens. A representation of HRED is given in Figure~\ref{hrnn_model}. 

In dialogue, the \textit{encoder} RNN maps each utterance to an utterance vector.
The utterance vector is the hidden state obtained after the last token of the utterance has been processed.
The higher-level \textit{context} RNN keeps track of past utterances by processing iteratively each utterance vector.
After processing utterance $U_m$, the hidden state of the context RNN represents a summary of the dialogue up to and including turn $m$, which is used to predict the next utterance $U_{m+1}$. This hidden state can be interpreted as the continuous-valued state of the dialogue system.
The next utterance prediction is performed by means of a \textit{decoder} RNN, which takes the hidden state of the context RNN and produces a probability distribution over the tokens in the next utterance. The decoder RNN is similar to the RNN language model~\cite{mikolov2010recurrent}, but with the important difference that the prediction is conditioned on the hidden state of the context RNN. It can be interpreted as the response generation module of the dialogue system.
The encoder, context and decoder RNNs all make use of the GRU hidden unit~\cite{Cho-et-al-EMNLP2014-better}. 
Every-where else we use the hyperbolic tangent as activation function. 
It is also possible to use the maxout activation function between the hidden state and the projected word embeddings of the decoder RNN \cite{goodfellow2013maxout}.
The same encoder RNN and decoder RNN parameters are used 
for {\nobreak every} utterance in a dialogue. This helps the model generalize across utterances.
Further details of the architecture are described by Sordoni et al.~\shortcite{sordoni2015ahier}.

For modeling dialogues, we expect the HRED model to be superior to the standard RNN model for two reasons. 
First, because the context RNN allows the model to represent a form of common ground between speakers,
e.g.\@ to represent topics and concepts shared between the speakers using a distributed vector representation,
which we hypothesize to be important for building an effective dialogue system \cite{clark1991grounding}.
Second, because the number of computational steps between utterances is reduced. 
This makes the objective function more stable w.r.t.\@ the model parameters, and helps propagate the training signal for first-order optimization methods
\cite{sordoni2015ahier}.


\subsection{Bidirectional HRED}
In HRED, the utterance representation is given by the last hidden state of the encoder RNN. This architecture worked well for web queries, but may be insufficient for dialogue utterances, which are longer and contain more syntactic articulations than web queries.
For long utterances, the last state of the encoder RNN may not reflect important information seen at the beginning of the utterance.
Thus, we also experiment with a model where the utterance encoder is a \emph{bidirectional} RNN.
Bidirectional RNNs run two chains: one forward through the utterance tokens and another backward, i.e.\@ reversing the tokens in the utterance. 
The forward hidden state at position $n$ summarizes tokens preceding position $n$ and the backwards hidden state summarizes tokens following position $n$.\footnote{The output of the bidirectional RNN is always based on the utterance before the current utterance of the decoder RNN.}
To obtain a fixed-length representation for the utterance, we summarize the information in the forward and backward RNN hidden states by either:
1) taking the concatenation of the last state of each RNN as input to the context RNN, or 2) applying $L_2$ pooling over the temporal dimension of each chain, and taking the concatenation of the two pooled states as input to the context RNN.\footnote{$L_2$ pooling over an utterance $U_m$ is defined as $\sqrt{1/N_m \sum_{n=1}^{N_m} h_n^2}$, where $h_n$ is the encoder RNN hidden state at position $n$, and $N_m$ is the length of the utterance.}
The bidirectional structure will effectively introduce additional short term dependencies, which has proven useful in similar architectures \cite{bahdanau2014neural,sutskever2014sequence}.
In  experiments below, we refer to this variant as HRED-Bidirectional.

\subsection{Bootstrapping from Word Embeddings and Subtitles Q-A}



The commonsense knowledge that the dialogue interlocutors share may be difficult to infer if the dataset is not sufficiently large. Therefore, 
our models may be improved by learning word embeddings from larger corpora. 
We choose to 
initialize our word embeddings $E$ with Word2Vec\footnote{
{\small \url{http://code.google.com/p/word2vec/}}}~\cite{mikolov2013distributedbetter} trained on the Google News dataset containing about 100 billion words. The sheer size of the
dataset ensures that the embeddings contain rich semantic information about each word. 

To learn a good initialization point for all model parameters, instead of only the word embeddings, 
we can further pretrain the model on a large non-dialogue corpus, which covers similar topics and types of interactions between interlocutors.
One such corpus is the Q-A \textit{SubTle} corpus containing about 5.5M Q-A pairs constructed from movie subtitles \cite{ameixa2014luke}.
We construct an artificial dialogue dataset by taking each $\{Q, A\}$ pair as a two-turn dialogue $D = \{U_1 = Q, U_2 = A\}$ and use this to pretrain the model.

\section{Dataset}

The \textit{MovieTriples} dataset has been developed by expanding and preprocessing the \textit{Movie-DiC} dataset by Banchs et al.~\shortcite{Banchs:2012:MMD:2390665.2390716} to make it fit the generative dialogue modeling framework. The dataset is available upon request.
Movie scripts span a wide range of topics,
contain long interactions with few participants 
and relatively few spelling mistakes and acronyms. 
As observed by Forchini~\shortcite{forchini2009spontaneity}:
\textit{``movie language can be regarded as a potential source for teaching and learning spoken language features''}.
Therefore, we believe bootstrapping a goal-driven spoken dialogue system based on movie scripts will improve performance.

We used the Python-based natural language toolkit NLTK \cite{BirdKleinLoper09} to perform tokenization and named-entity recognition.
All names and numbers were replaced with the \textit{\textless person\textgreater} and  \textit{\textless number\textgreater} tokens respectively \cite{Ritter:2010:UMT:1857999.1858019}.
To reduce data sparsity further, all tokens were transformed to lowercase
letters, and all but the 10,000 most frequent tokens were replaced with
a generic \textit{\textless unk\textgreater} token.


\begin{table}[tpb]
\renewcommand\arraystretch{1.1}
\small
    \begin{tabular}{ p{2.5cm} c  c  c }
     \midrule
        & \textbf{Training} & \textbf{Validation} & \textbf{Test} \\ \midrule
    Movies & 484 & 65 & 65 \\
    Triples & 196,308 & 24,717 &  24,271 \\
    Avg. tokens/triple & 53 & 53 & 55 \\
    Avg. unk/triple & 0.97 & 1.22 & 1.19 \\
    \bottomrule
    \end{tabular}
    \caption{\label{tab:stats}Statistics of the \textit{MovieTriples} dataset.}
\end{table}


We then generated ``triples'' $\{U_1, U_2, U_3\}$, i.e.~dialogues of three turns between two interlocutors A and B, for which A emits a first utterance $U_1$, B responds with $U_2$ and A responds with a last utterance $U_3$~\cite{sordoni2015aneural}.
To capture the interactive dialogue structure, a special end-of-utterance token is appended to all utterances and a continued-utterance token between breaks in lines from the same speaker. 
To avoid co-dependencies between triples coming from the same movie, we first split the movies into training, validation and test set, and then construct the triples. 
Statistics are reported in Table~\ref{tab:stats}.

\section{Experiments}

We evaluate the different variants of our HRED model, and compare against several alternatives, including basic $n$-gram models \cite{joshua2001bit}, a standard (non-hierarchical) RNN trained on the concatenation of the utterances in each triple, and a context-sensitive model (DCGM-I) recently proposed by Sordoni et al.~\shortcite{sordoni2015aneural}.

\subsection{Evaluation Metrics}

Accurate evaluation of a non-goal-driven dialogue system is an open problem \cite{GalleyBSJAQMGD15,pietquin2013survey,schatzmann2005quantitative}. There is no well-established method for automatic evaluation, and human-based evaluation is expensive. 
Nevertheless, for probabilistic language models word perplexity is a well-established performance metric \cite{Bengio-nnlm2003,mikolov2010recurrent}, and has been suggested for generative dialogue models previously \cite{pietquin2013survey}. We define word perplexity:
\begin{align} \label{eq:word_perplexity}
\exp \left ( - \dfrac{1}{N_W} \sum_{n=1}^N \log P_{\theta}(U_1^n, U_2^n, U_3^n) \right ),
\end{align}
for a model with parameters $\theta$, dataset with $N$ triples $\{U_1^n, U_2^n, U_3^n\}_{n=1}^N$, and $N_W$ the number of tokens in the entire dataset. Lower perplexity is indicative of a better model.  Perplexity explicitly measures the model's ability to account for the syntactic structure of the dialogue (e.g. turn-taking) and the syntactic structure of each utterance (e.g. punctuation marks). In dialogue, the distribution over the words in the next utterance is highly multi-modal, e.g.\@ there are many possible answers, which makes perplexity particularly appropriate because it will always measure the probability of regenerating the exact reference utterance.

We also consider the word classification error (also known as word error-rate). This is defined as the number of words in the dataset the model has predicted incorrectly divided by the total number of words in the dataset.\footnote{For a word prediction to be counted as correct, both the word and its position in the utterance must be correct.}
Each word contributes either zero or $1/N_W$ to the count, which means that it is more robust to unlikely (e.g.\@ unpredictable) words.
However, it is also less fine-grained than word perplexity. Instead of measuring the whole distribution, it only measures the regions of high probability. 

Ultimately, 
we care about generating syntactically and semantically coherent dialogues. For example, utterances which are grammatically correct and reflect the distribution of topics in the corpus, and whole dialogues which reflect the interaction patterns and topical evolutions of the dialogues in the corpus.
Despite having been proposed before \cite{pietquin2013survey,schatzmann2005quantitative}. it is not clear how well word perplexity and word classification errors correlate with this goal.
Nevertheless, optimizing probabilistic models using word perplexity has shown promising results in several machine learning tasks including statistical machine translation \cite{auli2013joint,sutskever2014sequence,bahdanau2014neural}, speech recognition \cite{hinton2012deep,deng2013machine} and image caption generation \cite{kiros2014multimodal,Vinyals_2015_CVPR}. Based on these empirical findings, we expect to be able to discriminate between models based on word perplexity, and to use word classification error and qualitative analysis of generated dialogues to understand the performance of the models in depth.

\begin{table*}
\renewcommand{\arraystretch}{1.1}
  \small
  \centering
    \begin{tabular}{| c | c | c | c | c | }
    \hline
    \textbf{Model} & \textbf{Perplexity} & \textbf{Perplexity@$\mathrm{\mathbf{U_3}}$} & \textbf{Error-Rate} & \textbf{Error-Rate@$\mathrm{\mathbf{U_3}}$} \\ \hline
    \parbox[t]{5.5cm}{Backoff N-Gram} & $64.89$ & $65.05$ & - & - \\ \hline
    \parbox[t]{5.5cm}{Modified Kneser-Ney} & $60.11$ & $54.75$ & - & - \\ \hline
    \parbox[t]{5.5cm}{Absolute Discounting N-Gram} & $56.98$ & $57.06$ & - & - \\ \hline
    \parbox[t]{5.5cm}{Witten-Bell Discounting N-Gram} & $53.30$ & $53.34$ & - & - \\ \hline
    \hline
    \parbox[t]{5.5cm}{RNN} & $35.63 \pm 0.16$ & $35.30 \pm 0.22$ & $66.34\% \pm 0.06$ & $66.32\% \pm 0.08$ \\ \hline
    \parbox[t]{5.5cm}{DCGM-I} & $36.10 \pm 0.17$ & $36.14 \pm 0.26$ & $66.44\% \pm 0.06$ & $66.57\% \pm 0.10$ \\ \hline
    \parbox[t]{5.5cm}{HRED} & $36.59 \pm 0.19$ & $36.26 \pm 0.29$ & $66.32\% \pm 0.06$ & $66.32\% \pm 0.11$ \\ \hline\hline
    \parbox[t]{5.5cm}{HRED + Word2Vec} & $33.95 \pm 0.16$ & $33.62 \pm 0.25$ & $66.06\% \pm 0.06$ & $66.05\% \pm 0.09$ \\ \hline\hline
    \parbox[t]{5.5cm}{RNN + SubTle} & $27.09 \pm 0.13$ & $26.67 \pm 0.19$ & $64.10\% \pm 0.06$ & $64.07\% \pm 0.10$ \\ \hline
    \parbox[t]{5.5cm}{HRED + SubTle} & $27.14 \pm 0.12$& $26.60 \pm 0.19$ & $64.10\% \pm 0.06$ & $64.03\% \pm 0.10$ \\ \hline
    \parbox[t]{5.5cm}{HRED-Bi.\@ + SubTle} & $\mathbf{26.81 \pm 0.11}$ & $\mathbf{26.31 \pm 0.19}$ & $\mathbf{63.93\% \pm 0.06}$ & $\mathbf{63.91\% \pm 0.09}$ \\
    \hline
    \end{tabular}
    \caption{Test set results computed on $\{U_1, U_2, U_3\}$ and solely on $\{U_3\}$ conditioned on $\{U_1, U_2\}$. Standard deviations are shown for all neural models. Best performances are marked in bold.}
\label{table:n_grams_vs_neural_models}
\end{table*}

\begin{table*}
\renewcommand{\arraystretch}{1.1}
  \small
  \centering
    \begin{tabular}{| p{7cm} | p{4.0cm} | p{4cm} | }
    \hline
    \textbf{Reference} ($\mathrm{\mathbf{U_1}}$, $\mathrm{\mathbf{U_2}}$) & \textbf{MAP} & \textbf{Target} ($\mathrm{\mathbf{U_3}}$) \\ \hline
    \parbox[t]{6cm}{$\mathrm{U_1}$: yeah , okay . \\ $\mathrm{U_2}$:  well , i guess i ' ll be going now .} & i ' ll see you tomorrow . & yeah . \\ \hline
    \parbox[t]{6cm}{$\mathrm{U_1}$: oh . \textless continued\_utterance\textgreater \ oh . \\ $\mathrm{U_2}$:  what ' s the matter , honey ?} & i don ' t know . & oh . \\ \hline
    \parbox[t]{6cm}{$\mathrm{U_1}$: it ' s the cheapest . \\ $\mathrm{U_2}$: then it ' s the worst kind ?} & no , it ' s not . & they ' re all good , sir . \\ \hline
    \parbox[t]{6cm}{$\mathrm{U_1}$: \textless person\textgreater \ ! what are you doing ? \\ $\mathrm{U_2}$: shut up ! c ' mon .} & what are you doing here ? & what are you that crazy ? \\ \hline
    \end{tabular}
    \caption{MAP outputs for HRED-Bidirectional bootstrapped from \textit{SubTle} corpus. The first column shows the reference utterances, where $U_1$ and $U_2$ are respectively the first and second utterance in the test triple. The second column shows the MAP output produced by beam-search conditioned on $U_1$ and $U_2$. The third column shows the actual third utterance in the test triple.}
\label{table:model_samples}
\end{table*}

\subsection{Training Procedure}

To train the neural network models, we optimized the log-likelihood of the triples using the recently proposed Adam optimizer~\cite{kingma2014adam}.\footnote{We truncated all triples to have a maximum size of $80$ tokens, and could therefore apply backpropagation on the full token sequences.}
Our implementation relied on the open-source Python library Theano~\cite{Better-Bastien-Theano-2012}.\footnote{The model implementations can be found on GitHub: \\ \url{https://github.com/julianser/hed-dlg} \\ \url{https://github.com/julianser/rnn-lm}.}
The best hyperparameters of the models were chosen by early stopping with patience on the validation set perplexity~\cite{bengio2012practical}.
We initialized the recurrent parameter matrices as orthogonal matrices, and all other parameters from a Gaussian random distribution with mean zero and standard deviation $0.01$.
For the baseline RNN, we tested hidden state spaces $d_h = 200$, $300$ and $400$. 
For HRED we experimented with encoder and decoder hidden state spaces of size $200$, $300$ and $400$. 
Based on preliminary results and due to GPU memory limitations, we limited ourselves to size $300$ when not bootstrapping or bootstrapping from Word2Vec, and to size $400$ when bootstrapping from \textit{SubTle}.
Preliminary experiments showed that the context RNN state space at and above $300$ performed
similarly, so we fixed it at $300$ when not bootstrapping or bootstrapping from Word2Vec, 
and to $1200$ when bootstrapping from \textit{SubTle}.
For all models, we used word embedding of size $400$ when bootstrapping from \textit{SubTle} and of size $300$ otherwise.
To help generalization, we used the maxout activation function, between the hidden state and the projected word embeddings of the decoder RNN, when not bootstrapping and when bootstrapping from Word2Vec. 
We used $L_2$ pooling for all HRED models, except when bootstrapping from \textit{SubTle} since it appeared to perform slightly worse.

\paragraph{Bootstrapping Word Embeddings} Our embedding matrix $E$ was initialized using the publicly available 300 dim.\@ Word2Vec embeddings trained on the Google News corpus~\cite{mikolov2013distributedbetter}. Special dialogue tokens, which did not exist in the Word2Vec embeddings, were initialized from a Gaussian random distribution as before. 
The training procedure was carried out in two stages.
First, we trained each neural model with fixed Word2Vec embeddings.
During this stage, we also trained the speech act and placeholder tokens, 
together with tokens not covered by the original Word2Vec embeddings. 
In the second stage, we then trained all parameters of 
each neural model until convergence.
\paragraph{Bootstrapping SubTle}
We processed the \textit{SubTle} corpus by following the same procedure as used for \textit{MovieTriples}, but treating the last utterances $U_3$ as empty. The final \textit{SubTle} corpus contained 5,503,741 Q-A pairs, and a total of 93,320,500 tokens.
When bootstrapping from the \textit{SubTle} corpus, we found that all models performed slightly better when randomly initializing and learning the word embeddings from \textit{SubTle} compared to fixing the word embeddings to those given by Word2Vec. 
For this reason, we do not report results combining bootstrapping from the \textit{SubTle} corpus with Word2Vec word embeddings. 

The HRED models were pretrained for approximately four epochs on the 
\textit{SubTle} dataset, after which performance did not appear to improve further. 
Then, we fine-tuned the pretrained models on the \textit{MovieTriples} dataset 
holding the word embeddings fixed. 

\subsection{Empirical Results} 

Our results are presented in Table~\ref{table:n_grams_vs_neural_models}. All neural models beat state-of-the-art $n$-grams models substantially w.r.t.\@ both word perplexity and word classification error.
Without bootstrapping, the RNN model performs similarly to the more complex DCGM-I and HRED models. This can be explained by the size of the dataset, which makes it easy for the HRED and DCGM-I model to overfit.
The last four lines of Table~\ref{table:n_grams_vs_neural_models} confirm that bootstrapping the model parameters achieves significant gains in both measures. Bootstrapping from \textit{SubTle} is particularly useful since it allows a gain of nearly 10 perplexity points compared to the HRED model without bootstrapping. We believe that this is because it trains all model parameters, unlike bootstrapping from Word2Vec, which only trains the word embeddings.

In general, we find that the gains due to architectural choice are smaller than those obtained by bootstrapping, which can be explained by the fact that we are in a regime of relatively little training data compared to other natural language processing tasks, such as machine translation, and hence we would expect the differences to grow with more training data and longer dialogues. 
Overall, the bidirectional structure appears to capture and retain information from the $U_1$ and $U_2$ utterances better than either the RNN and the original HRED model. This confirms our earlier hypothesis, and demonstrates the potential of HRED for modeling long dialogues.

\subsection{MAP Outputs}
\newcommand{\argmax}{\operatornamewithlimits{argmax}}
%
%
We also considered the use of beam-search for RNNs~\cite{graves2012seq} to approximate
the most probable (MAP) last utterance, $U_3$, given the first two
utterances, $U_1$ and $U_2$.
MAP outputs are presented in Table \ref{table:model_samples} for HRED-Bidirectional bootstrapped from \textit{SubTle} corpus. As shown here, the model often produced sensible answers. However, in fact, the majority of the predictions were generic, such as \textit{I don't know} or \textit{I'm sorry}.  We observed the same phenomenon for the RNN model, and similar observations can be inferred by remarks in the recent literature \cite{sordoni2015aneural,vinyals2015neural}. 
To the best of our knowledge, we are the first to emphasize and discuss it in details.\footnote{After publishing the first draft of this paper, Li et al.\@ \shortcite{DBLP:journals/corr/LiGBGD15} investigated this problem further and proposed to change the objective function at test time to also maximize the mutual information between the generated utterance and the previous utterances.}

There are several possible explanations for this behavior. 
Firstly, due to data scarcity, the models may only have learned to predict the most frequent utterances. 
Since the dialogues are inherently ambiguous and multi-modal, predicting them accurately would require more data than other natural language processing tasks. 
Secondly, the majority of tokens were punctuation marks and pronouns. Since every token is weighted equally during training, the gradient signal of the neural networks is dominated by these punctuation and pronoun tokens. This makes it hard for the neural networks to learn topic-specific embeddings and even harder to predict diverse utterances. 
This suggests exploring neural architectures which explicitly separate semantic structure from syntactic structure. 
Finally, the context of a triple may be too short. In that case, the models should benefit from longer contexts and by conditioning on other information sources, such as semantic and visual information.

An important implication of this observation is that metrics based on MAP outputs (e.g.\@ cosine similarity, BLEU, Levenshtein distance) will primarily favor models that output the same number of punctuation marks and pronouns as are in the test utterances, as opposed to similar semantic content (e.g. nouns and verbs). This would be systematically biased and not necessarily in any way correlate with the objective of producing appropriate responses. We therefore cannot justify the use of such metrics when the results are known to lack diversity.

Nevertheless, we also note that this problem did not occur when we generated stochastic samples (as opposed to the MAP outputs). In fact, the stochastic samples contained a large variety of topic-specific words and often appeared to maintain the topic of the conversation.

\section{Conclusion and Future Work}
We have demonstrated that a hierarchical recurrent neural network generative
model can outperform both $n$-gram based models and baseline neural network models on the task of modeling utterances and speech acts.
 To support our investigation, we introduced a novel dataset called {\nobreak \textit{MovieTriples}} based on movie scripts, 
which are suitable for modeling long, open domain dialogues close to human spoken language.
In addition to the recurrent hierarchical architecture, we found two crucial ingredients for improving performance: 
the use of a large external monologue corpus to initialize the word embeddings, and the use of a 
large related, but non-dialogue, corpus in order to {\em pretrain} the recurrent net. 
This points to the need for larger dialogue datasets.  

Empirical performance of all models was measured using perplexity. While this is an established measure for generative models, in the dialogue setting utterances may be overwhelmed by many common words especially {\nobreak arising} from colloquial or informal exchanges.  It may be fruitful to investigate other measures of performance for generative dialogue systems.
We also considered actual responses produced by the model. The MAP outputs tended to produce somewhat generic, but conversationally acceptable, responses. Stochastic samples from the model produced more diverse dialogues.

Future work should study models for full length dialogues, as opposed to triples,
and model other speech acts, such as interlocutors entering or leaving the dialogue 
and executing actions.
Finally, our analysis of the model MAP outputs suggests that it would be beneficial to
include longer and additional context, including other modalities such as audio and video.


\newpage
\section{Acknowledgements}
The authors acknowledge IBM Research, NSERC, Canada Research Chairs, Nuance Foundation, CIFAR and Compute
Canada for funding and resources. The authors thank Ryan Lowe, Laurent Charlin and Nissan Pow
for constructive feedback. The authors thank Rafael E.\@ Banchs for providing the \textit{Movie-DiC} dataset, and Luisa Coheur for providing the \textit{SubTle} dataset. The authors also thank the anonymous AAAI reviewers for their helpful feedback.



\fontsize{9.5pt}{10.5pt}
\bibliographystyle{aaai}
\bibliography{strings,ml,bib}

\end{document}